\documentclass[conference]{IEEEtran}
\IEEEoverridecommandlockouts
\usepackage{balance}
\usepackage{url}
\usepackage[hidelinks,breaklinks]{hyperref}
\usepackage{cite}
\usepackage{amsmath,amssymb,amsfonts}
\usepackage{algorithmic}
\usepackage{graphicx}
\usepackage{textcomp}
\usepackage{xcolor}
\def\BibTeX{{\rm B\kern-.05em{\sc i\kern-.025em b}\kern-.08em
    T\kern-.1667em\lower.7ex\hbox{E}\kern-.125emX}}
\begin{document}

\title{A Modular Vision-Language-Action Robotics Framework for Indoor Environments}

\author{\IEEEauthorblockN{Anindya Jana, Snehasis Banerjee, Arup Sadhu, Ranjan Dasgupta}
\IEEEauthorblockA{\textit{Physical AI Research} \\
\textit{TCS Research, Tata Consultancy Services}\\
Kolkata, India \\
\{firstname.lastname\}@tcs.com}
}

\maketitle

\begin{abstract}
This paper presents an integrated system for the CMU Vision-Language-Action (VLA) Challenge, designed to enable an autonomous agent to perform complex tasks based on natural language instructions. Our framework employs a modular architecture that orchestrates environment mapping, question processing, and navigation. The system operates in two parallel streams: a perception pipeline that constructs a semantic voxel map from real-time camera feeds using OwlViT embeddings, and a language pipeline that classifies user commands with a Vision-Language Model (Gemini). The mapping is time-constrained; the system proceeds with a partial map if a 500-second exploration limit is reached. The classified query is then grounded in the geometric and semantic context of the map to generate a detailed prompt for the VLM. This yields an actionable output, demonstrating a capable solution for bridging the gap between human language and robotic action.
\end{abstract}


\section{Introduction}
The ability for autonomous systems to understand and act upon high-level, natural language instructions from humans is a cornerstone of next-generation robotics. This capability requires a deep integration of perception, language understanding, and action generation. The CMU Vision-Language-Action (VLA) Challenge\cite{b1} provides a benchmark for this problem, tasking systems with navigating complex indoor environments and responding to a variety of user queries.

Our goal is to develop a robust and modular framework that can systematically parse linguistic commands, ground them within a rich, semantic representation of the environment, and execute the corresponding tasks. Traditional approaches often struggle with the ambiguity and variability of human language. To overcome this, we leverage the reasoning capabilities of modern Vision-Language Models (VLMs)\cite{b17} in conjunction with a precise, metric-semantic map of the agent's surroundings.

This paper details our system's architecture, which is built upon the Robot Operating System (ROS)\cite{b10} . Our primary contributions are: 1) A semantic voxel map that stores not only geometric information but also rich feature embeddings for detected objects using the OwlViT model \cite{b9}, built under a strict time limit. 2) A novel, VLM-driven pipeline that first classifies the user's intent and then generates a context-aware prompt by fusing the query with real-time data from the semantic map. 3) A state-driven execution flow that manages exploration, query handling, and navigation, creating a complete loop from instruction to action.

\section{About the CMU VLA Challenge}
\begin{figure}
    \centering
    \includegraphics[width=1\linewidth]{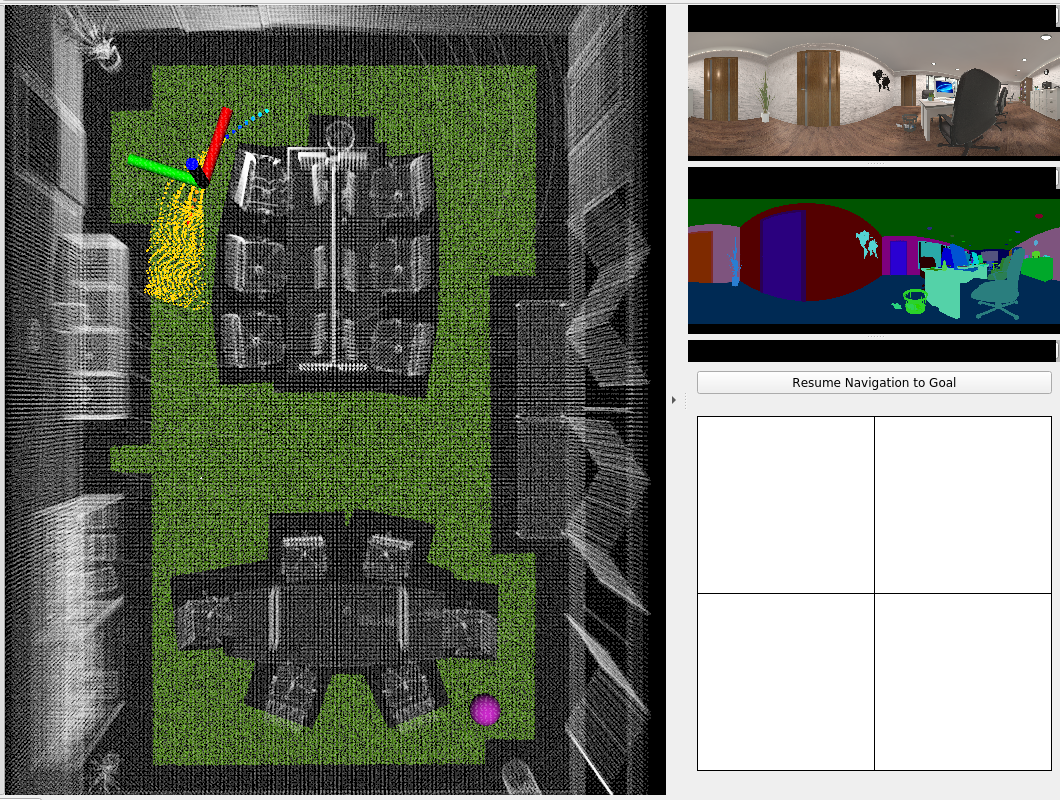}
    \caption{CMU-VLA-Challenge Setup with a given example scene in Rviz.}
    \label{fig:placeholder}
\end{figure}

The CMU VLA Challenge \cite{b1} is designed to advance the field of embodied AI by presenting a set of complex tasks that require an agent to seamlessly integrate navigation, perception, and natural language understanding. The challenge takes place in simulated indoor environments, specifically leveraging AI Habitat \cite{b11} and Unity, which provide realistic physics and high-fidelity 3D scenes based on Matterport3D models \cite{b6}. A total of 18 distinct Unity scenes, categorized as offices, apartments, and houses, are used for the challenge, providing a diverse range of layouts. For the development and demonstration of our system, we utilized one of the office environments provided by the challenge, as depicted in  Fig.~\ref{fig:placeholder}.

The core task is for an autonomous agent to explore an unknown environment and answer a series of questions based on a single language command. These questions fall into three distinct categories, testing a range of the agent's capabilities:
\begin{itemize}
    \item \textbf{Numerical:} These questions require a quantitative answer about the environment, compelling the agent to perform comprehensive exploration and object counting (e.g., How many photos are on the TV cabinet?").
    \item \textbf{Visualization/Object Reference:} These commands task the agent with locating a specific object in the scene, which often involves understanding spatial prepositions and attributes (e.g., ``Find the sofa."). The expected output is typically the object's coordinates or a bounding box.
    \item \textbf{Waypoint-Based Navigation:} These are complex instructions that require the agent to generate and follow a path defined by object landmarks (e.g., ``Go to the kitchen area, then find the nearest table."). This tests the agent's ability to translate high-level commands into low-level motor actions.
\end{itemize}

For each language command, the system is relaunched in a new, previously unseen scene. This ensures that no information from previous explorations is retained, forcing the agent's  model to be built from scratch for every task.

Furthermore, each task has a total time limit of 10 minutes. Within this window, the agent must perform all necessary actions—exploration, mapping, reasoning, and navigation—to arrive at and deliver its final response. 

\section{Building a VLA Model: Initial Approach}
Our initial development approach was to create a simple architecture that did not rely on an explicit world model. We replaced the dummy model provided by the challenge with a vision-language model, \texttt{blip2-opt-2.7b} \cite{b12}, from Hugging Face . This model processed the agent's live camera feed to answer questions based solely on its immediate, current view. For example, when asked ``How many chairs are in the room?", the model would respond based on the number of chairs visible in that single frame.

The primary limitation of this approach was its complete lack of environmental memory or spatial awareness. It could not answer questions about objects outside its field of view or reason about the global layout of the scene. To address this, we implemented a basic exploration strategy where the agent moved in random directions while using the BLIP2 model to detect and log the names and coordinates of objects into a simple text file.

However, this method proved to be highly inefficient. The random exploration was unstructured, time-consuming, and often resulted in the agent repeatedly scanning the same areas while missing entire rooms. The text-file ``memory" was difficult to query for spatial relationships, and the overall approach was too slow and unreliable to meet the challenge's 10-minute time limit. This failure highlighted the absolute necessity of building a structured, persistent world model for effective task completion.

\section{A Structured World Model Approach (VLA-3D Inspired)}
Recognizing the limitations of a memoryless agent, our design shifted towards an approach centered on building a comprehensive, structured representation of the environment. Our inspiration was drawn from data-generation pipelines like VLA-3D\cite{b3}, which leverage detailed 3D scan data to create  multi-modal scene descriptions. This methodology consists of three conceptual stages:
\begin{enumerate}
    \item \textbf{3D Scan Processing:} Utilizing the provided 3D environment models, we process the underlying point cloud data to establish a foundational geometric understanding of the scene.
    \item \textbf{Scene Graph Generation:} From the geometric data, a structured scene graph is generated. This abstracts the environment into a set of nodes (objects) and edges (their spatial and semantic relationships, such as \textit{a cup is on a table}).
    \item \textbf{Language Generation:} Finally, the structured scene graph is traversed to generate rich, descriptive language that captures the state of the environment\cite{b16}, which can then be used by an LLM to answer questions.
\end{enumerate}
During this phase, we experimented with some VLA model, which showed  performance in answering spatial and numerical questions when provided with this kind of structured input. This shift from a purely reactive model to a world-model-based paradigm was the pivotal step that informed our final, successful architecture. We ultimately adapted this concept from using a pre-generated scene graph to dynamically building a semantic map during exploration.

% Using a single-column figure for better placement
\begin{figure}[t]
\centering
\includegraphics[width=\linewidth]{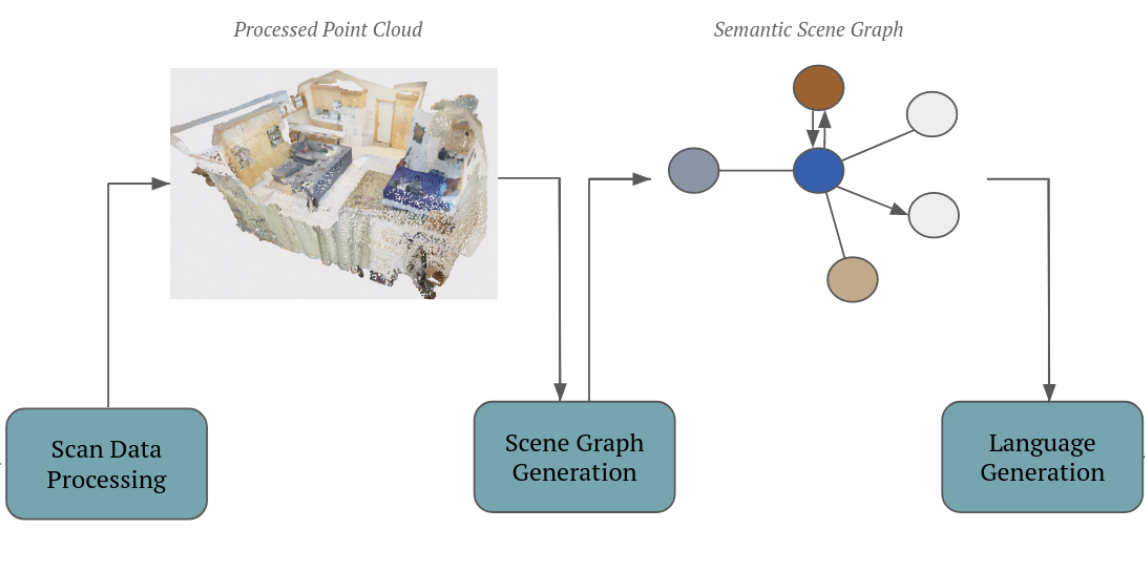} 
\caption{The pipeline for processing a raw 3D scan into a structured semantic scene graph and generating a corresponding referential language statement.}

\label{fig:architecture}
\end{figure}

\subsection{Limitations for Real-World Application}
While the VLA-3D\cite{b3} inspired pipeline provided a powerful conceptual framework, its primary prerequisite is the availability of a pre-existing 3D model of the environment. Since the robot will be tested in a real world scenario,there will be no 3D model available. This is a significant limitation for a truly autonomous agent, as such models are not available in novel, real-world scenarios. An agent must be able to build its understanding of the world purely from its own sensor data during exploration, not from a perfect, pre-supplied map.

Therefore, this realization was the key factor that informed our final architecture. We adapted the core concept: instead of relying on a pre-generated scene graph, our final system dynamically builds its own semantic map from sensor readings in real-time. This retains the benefit of a structured world model while removing the dependency on prior environmental data, making it a viable approach for the challenge and for potential real-world deployment.

\section{A Comparative Framework: SORT3D}
To provide context for our work, it is useful to examine other contemporary modular frameworks designed for similar tasks. One such system is SORT3D \cite{b2}, a zero-shot pipeline for 3D referential grounding\cite{b7} that decomposes the task into a series of sub-problems. The goal of SORT3D is to identify a specific object in a 3D scene based on a free-form natural language expression (e.g., "the mug on the table near the window"). Its methodology can be broken down into four main components.
\begin{enumerate}
\item \textbf{Instance-level Semantic Mapping:} For real-world deployment, SORT3D\cite{b2} uses a semantic mapping module that fuses data from a 360° camera and a 3D LiDAR to detect objects and generate their 3D bounding boxes. However, for its benchmark evaluations on datasets like ReferIt3D\cite{b13}, the framework utilizes \textbf{ground truth bounding boxes and instance segmentations}.

\item \textbf{Enhancing Perception with 2D Captions:} Recognizing that pure 3D perception often fails to capture fine-grained object attributes, SORT3D uses modern vision language models (VLMs) to enrich its understanding. For each detected object, it generates descriptive captions from 2D image \cite{b15}. A VLM is prompted to describe the object's color, material, shape, and other attributes, providing a much richer set of semantic information than a simple class label.

\item \textbf{Filtering for Relevant Objects:} Indoor environments can contain hundreds of objects, most of which are irrelevant to a given command. To manage this complexity, SORT3D employs an LLM-based filtering module. Given a command like "Find the nightstand to the right of the bed," this module first extracts the key object nouns ("nightstand," "bed") and then filters the full scene list to include only instances of these objects, dramatically narrowing the search space for the final step.

\item \textbf{LLM-based Spatial Reasoning:} Finally, the filtered list of relevant objects, along with their rich captions and 3D coordinates, is passed to a Large Language Model. This LLM is augmented with a spatial reasoning toolbox to interpret the spatial relationships described in the command (e.g., "to the right of") and identify the single target object being referenced.
\end{enumerate}

% Use figure* to span both columns
\begin{figure*}
\includegraphics[width=0.9\textwidth]{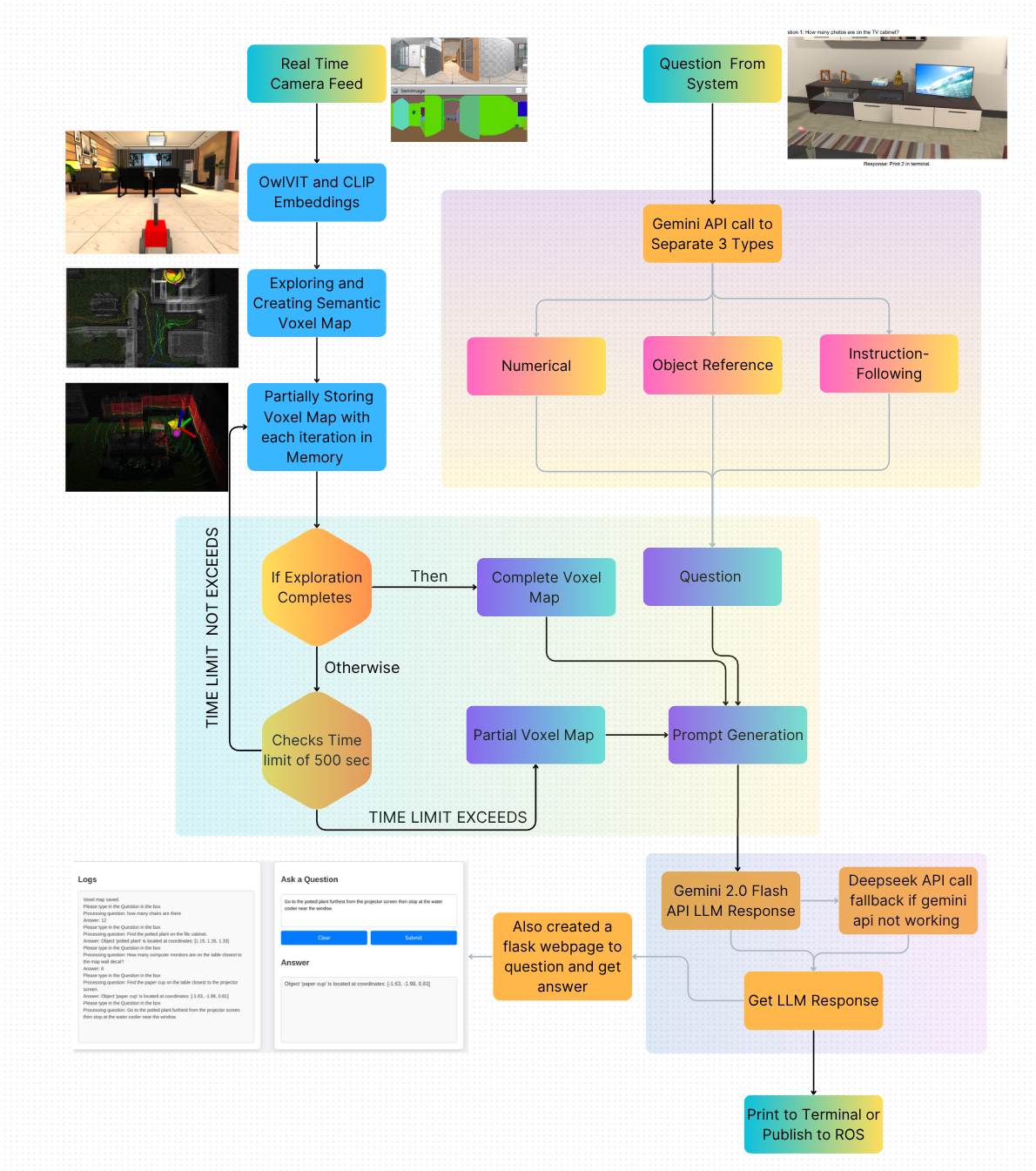} 
\caption{System architecture diagram illustrating the parallel pipelines for mapping and query processing. The system takes real-time camera feeds and user questions, builds a time-limited semantic voxel map, classifies the query, grounds it in the map's context to generate a prompt, and uses an LLM to produce the final actionable response.}
\label{fig:Screenshot from 2025-09-16 15-43-15}
\end{figure*}

\section{Our Approach: System Architecture and Implementation}

Our system is implemented as a collection of interconnected nodes within the Robot Operating System (ROS)\cite{b10} framework. The entire process, from environmental perception to task execution, is automated and does not require manual intervention, a key distinction from frameworks that rely on pre-existing models or manual navigation for semantic data collection. The implementation can be divided into four primary stages: Autonomous Exploration\cite{b8}, Semantic Mapping\cite{b18}, VLM-Powered Query Processing, and Action Generation. The complete architecture, from sensor input to action output, is depicted in Fig.~\ref{fig:Screenshot from 2025-09-16 15-43-15}.

\subsection{Autonomous Exploration and 2D Grid Mapping}

The foundation of our system is its ability to autonomously explore an unknown environment. This process begins with a grid map handling module, which takes input  from a LiDAR sensor and dynamically constructs a 2D  grid map\cite{b19}. This map serves as the basis for all navigation tasks.

Once the 2D map is generated, the planner node initiates a full-coverage exploration strategy. Rather than relying on random movements, the planner systematically generates a set of waypoints to ensure thorough coverage. This is achieved by:
\begin{enumerate}
    \item \textbf{Identifying Traversable Points:} The planner samples points across the grid, ensuring they are a minimum distance apart to create an efficient graph of visitable locations. A distance transform method is used to ensure these points maintain a safe buffer from walls and obstacles.
    \item \textbf{Optimal Path Calculation:} The planner treats the visitable points as a network of connected locations. It then calculates the most efficient route that visits every single point, creating an optimal tour that minimizes travel time and avoids unnecessary backtracking.
    \item \textbf{Waypoint Execution:} The final sequence of waypoints is passed to a dedicated waypoint controller, a navigation node that subscribes to odometry data and publishes commands to drive the robot to each waypoint in sequence, thereby completing the autonomous exploration phase.
\end{enumerate}

\subsection{Semantic Voxel Map Construction}

Simultaneously with the 2D navigation mapping, a richer, 3D semantic map is constructed by the semantic mapping module\cite{b18}. This module listens for  messages published by an upstream object detection node. For each detected object marker, the system:

\begin{enumerate}
    \item \textbf{Perception and Embedding:} The process begins with the real-time camera feed from the agent. This visual data is processed to detect objects and generate high-dimensional feature embeddings for them using the google/owlvit-base-patch32 model\cite{b9}. These embeddings capture the rich semantic meaning of each object.
    \item \textbf{Iterative Voxel Map Construction:} As the agent explores its environment, these semantic embeddings, along with their corresponding 3D coordinates and scale, are progressively integrated into a semantic voxel map\cite{b20}. This map is built iteratively, meaning it is partially stored and updated with each new perception cycle, allowing the system to have an immediate, albeit incomplete, world representation at any time.
    \item \textbf{Time-Constrained Exploration:} To ensure the system remains responsive, the exploration and mapping phase is constrained by a 500-second time limit. If the agent successfully completes its full-coverage navigation plan within this period, a ``Complete Voxel Map"\cite{b20} is finalized. However, if the time limit is exceeded, the exploration is halted, and the system proceeds with the ``Partial Voxel Map" constructed up to that point. This ensures the system can answer questions even if a full exploration is not feasible.
\end{enumerate}

\subsection{VLM-Powered Query Processing}

With the exploration complete and the semantic map built, a central question handling module orchestrates the language understanding pipeline.

\begin{enumerate}
    \item \textbf{Query Classification:} The user's natural language instruction is first sent to a (Gemini 2.0 Flash) which is designed to work as a VLM\cite{b17} with a carefully engineered prompt. The VLM classifies the instruction into one of three categories:

    \begin{itemize}
    \item \textbf{Numerical:} Questions about the quantity of objects (e.g., ``How many chairs are there?").
    \item \textbf{Object Reference:} Commands to locate a specific object (e.g., ``Find the potted plant near the table.").
    \item \textbf{Instruction-Following:} Commands that specify a path or trajectory (e.g., ``Go to the fridge, avoiding the sofa.").
\end{itemize}

    \item \textbf{Object Grounding\cite{b15}:} The system then uses the VLM again to extract the key object nouns from the instruction. For each noun, the localization module is invoked. This module loads the saved voxel map\cite{b20} and computes a text embedding for the query noun. By calculating the  similarity between the query embedding and all feature embeddings stored in the voxel map, it identifies the 3D points corresponding to the object(s) of interest.
    \item \textbf{Contextual Prompt Generation:} The retrieved information including object names, their 3D coordinates, scale, and their corresponding 2D grid map locations is formatted into a detailed, context-rich prompt. This crucial step grounds the abstract language query in the geometric and semantic reality of the robot's world model.
\end{enumerate}

\subsection{Post-Processing and Action Generation}

The final context-rich prompt is sent to the Gemini VLM for reasoning. The model's textual response is then parsed by a post-processing module, which triggers different actions based on the initial query classification.

\begin{itemize}
    \item \textbf{For Numerical Queries:} The response is parsed to extract the final integer answer, which is returned to the user.
    \item \textbf{For Object Reference Queries:} The module parses the grid coordinates from the VLM's answer. It then cross-references these with the localized points to find the object's global 3D coordinates and scale. This information is used to publish a \texttt{Marker} in RViz\cite{b21}, visually highlighting the identified object in the map.
    \item \textbf{For Instruction-Following Queries:} This triggers the most complex workflow. The VLM's response, a high-level plan (e.g., \texttt{Goto(x,y)}, \texttt{Avoidbetween((x1,y1),(x2,y2))}), is parsed. The \texttt{Avoidbetween} command dynamically modifies the 2D grid by marking the specified region as non-traversable. The system then uses the pathfinding algorithm to generate a smooth, direct path that connects the waypoints derived from the VLM's plan. This final path is converted to global coordinates and executed by the waypoint controller, enabling the robot to follow complex, language-defined trajectories.
\end{itemize}

\subsection{System Optimizations for Time-Constrained Exploration}
Given the strict 10-minute time limit of the CMU VLA Challenge\cite{b1}, initial trials revealed that the autonomous exploration and mapping phase for the office scene (Fig.~\ref{fig:placeholder}) consumed approximately 8 minutes and 42 seconds given the simulation was capped at 30 fps. Decreasing the time would give a higher score in challenge evaluation. To address this critical bottleneck, we implemented a series of parameter optimizations aimed at drastically reducing exploration time without compromising the system's question-answering accuracy. These changes successfully reduced the exploration phase to 4 minutes and 17 seconds, a 50\% reduction in time.

The optimizations were focused on two key areas: the navigation and planning module, and the semantic map construction process.

\subsubsection{Navigation and Planning Efficiency}
The primary goal here was to generate a shorter, more efficient exploration path and to allow the robot to traverse it more quickly.
\begin{itemize}
    \item \textbf{Increased Waypoint Spacing:} The minimum gap between sampled points in the full-coverage planner was increased from 0.6 meters to 1.25 meters. This creates a sparser coverage graph, reducing the total number of waypoints the robot must visit while still ensuring comprehensive environmental coverage.
    
    \item \textbf{Relaxed Waypoint Arrival Condition:} The arrival threshold for reaching a waypoint was increased from 0.65 meters to 1.0 meter. This allows the robot to consider a waypoint "reached" from a greater distance, minimizing time spent on fine-grained final positioning and enabling a more fluid transition to the next target.
\end{itemize}

\subsubsection{Semantic Map Construction Speed}
To complement the faster physical exploration, we also accelerated the process of building the semantic voxel map\cite{b20}.
\begin{itemize}
    \item \textbf{Increased Voxel Size:} The resolution of the voxel map was changed by increasing the voxel size to 0.15 meters. Larger voxels result in a lower-resolution but computationally lighter world model. This change significantly accelerates the process of adding new points and updating semantic features, as fewer total voxels need to be created and managed. While this reduces the spatial precision of the map, the 0.15m resolution proved sufficient for accurately grounding the object-level queries required by the challenge.
\end{itemize}

\section{Implementation Details}

\subsection{Models}
The system is built on ROS (Robot Operating System)\cite{b10} for modularity and communication between components. The deep learning components utilize PyTorch. We employ several key pre-trained models:
\begin{itemize}
    \item \textbf{OwlViT (\texttt{google/owlvit-base-patch32}):} Used for generating powerful CLIP embeddings \cite{b4} for both detected objects during mapping and text queries during localization. During our development, we also evaluated more recent vision-language models such as SigLip2 \cite{b14}. However, for our specific application requiring rapid processing within a time-constrained exploration phase, the CLIP embeddings from OwlViT\cite{b9} demonstrated greater computational efficiency. 
    \item \textbf{Gemini (\texttt{gemini-2.0-flash}):} Our primary high-level reasoning engine is Gemini 2.0 Flash\cite{b5}. We specifically chose the 'Flash' variant over more powerful models in the Gemini family due to its significantly lower latency. We used gemini-2.0-flash instead of gemini-2.5-flash because it showed much faster response.  In a time-critical robotics application where the system must explore and respond within a 10-minute window, minimizing the delay from API calls is very important. To further increase system reliability, Deepseek v3 is employed as a fallback API in case of failures with the primary service.
\end{itemize}

\subsection{User Interface and System Output}
To facilitate interaction with the system, a user-facing web interface was developed using Flask and Socket.IO. As shown in Fig.~\ref{fig:ui}.

This interface allows for the testing and demonstration of the system's ability to handle the three distinct question types required by the challenge. Below are some examples of question-and-answer pairs processed by our system, taken from the logs:

\begin{figure}[t]
\centering
\includegraphics[width=\linewidth]{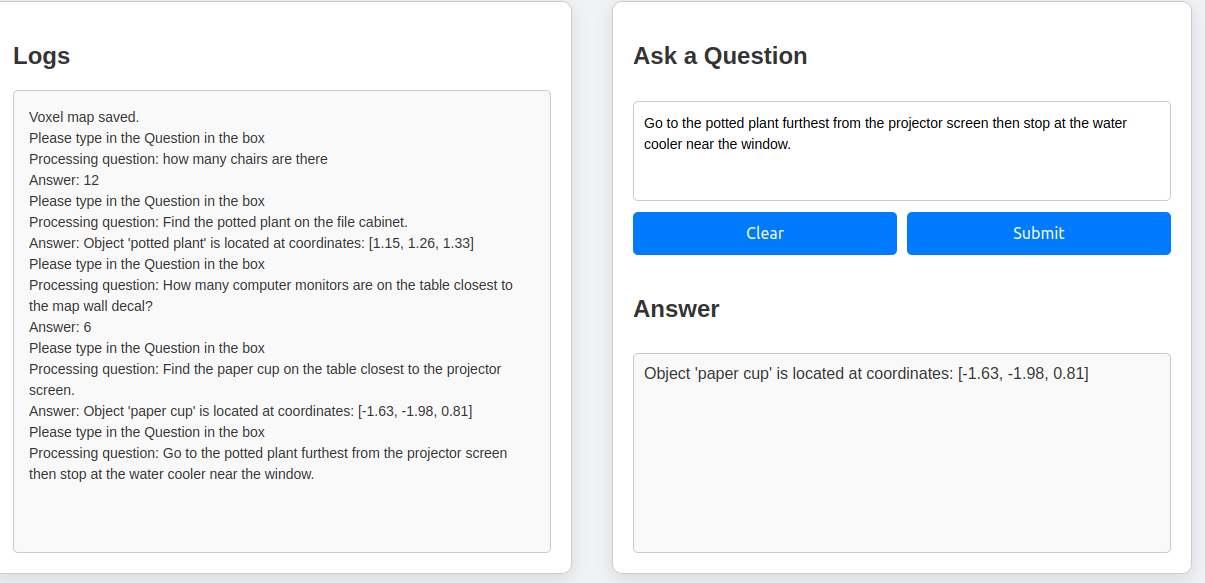} 
\caption{The Flask-based web user interface. The left panel shows a running log of system operations, including questions and answers. The right panel allows the user to submit a new natural language command and displays the processed result.}
\label{fig:ui}
\end{figure}

\begin{itemize}
    \item \textbf{Numerical Question:}
    \begin{itemize}
        \item \textit{Question:} ``how many chairs are there"
        \item \textit{Answer:} \texttt{12}
    \end{itemize}
    \vspace{0.5em} % Adds a little space between items
    \item \textbf{Object Reference Question:}
    \begin{itemize}
        \item \textit{Question:} ``Find the potted plant on the file cabinet."
        \item \textit{Answer:} \texttt{Object 'potted plant' is located at coordinates: [1.15, 1.26, 1.33]}
    \end{itemize}
    \vspace{0.5em}
    \item \textbf{Complex Object Reference:}
     \begin{itemize}
        \item \textit{Question:} ``Find the paper cup on the table closest to the projector screen."
        \item \textit{Answer:} \texttt{Object 'paper cup' is located at coordinates: [-1.63, -1.98, 0.81]}
    \end{itemize}
\end{itemize}
For instruction-following questions, such as ``Go to the potted plant...", the textual output confirms the target, and the primary output is the publication of waypoints to the ROS navigation stack.

\section{Limitations}
While our framework demonstrates a robust and integrated approach to the VLA challenge, we acknowledge several limitations that provide clear avenues for future research.

\begin{itemize}
    \item \textbf{Reliance on Online VLM APIs:} The current system depends on internet connectivity to make API calls to Gemini for high-level reasoning and classification. This reliance introduces latency and makes the system vulnerable to network failures, limiting its viability for deployment in environments without reliable internet access.

    \item \textbf{Assumption of a Static Environment:} The current architecture follows a sequential "explore, map, then act" perspective. The semantic map is constructed during the initial exploration phase and is then treated as a static representation of the world. This approach is not suitable for dynamic environments where objects or furniture might be moved after the mapping is complete.
    
\end{itemize}

\section{Conclusion}
We have presented a modular and effective framework for the CMU VLA Challenge\cite{b1} that successfully integrates semantic mapping, natural language understanding, and autonomous action. By combining the semantic feature extraction capabilities of OwlViT\cite{b9} with the advanced reasoning of large language models like Gemini, our system can interpret and execute a diverse range of commands. The three-stage query processing pipeline: classification, ground, and response, coupled with a time-constrained mapping strategy, proves to be a robust paradigm for solving complex VLA tasks.

Future work could focus on reducing the latency of VLM API calls, potentially by using smaller, locally-hosted models for simpler queries. Additionally, integrating a more sophisticated motion planner could allow for the execution of more complex and nuanced instructions, further closing the gap between human language and robotic autonomy.

\section*{Acknowledgment}
The authors would like to thank the organizers of the CMU VLA Challenge\cite{b1} for providing a stimulating and well-defined problem domain for advancing research in vision-language-action systems.
\balance

\end{document}